\documentclass[letterpaper, 10 pt, conference]{ieeeconf}  

\usepackage{algs}
\usepackage{amsfonts}
\usepackage{amsmath}
\usepackage{times}
\usepackage[numbers]{natbib}
\usepackage{multicol}
\usepackage[bookmarks=true]{hyperref}
\usepackage{lipsum}
\usepackage{comment}
\usepackage{balance}
\usepackage{graphicx}
\usepackage{xcolor}
\usepackage{listings}
\usepackage{tabularx}
\usepackage[linesnumbered,ruled,noend,vlined]{algorithm2e}

\SetCommentSty{mycommfont}

\usepackage{seqsplit}

\usepackage[frozencache,cachedir=.]{minted}
\definecolor{bg}{rgb}{0.95, 0.95, 0.92}

\usepackage{caption}
\IEEEoverridecommandlockouts                              

\title{\Large \bf
ad-trait: A Fast and Flexible Automatic Differentiation Library in Rust
}

\author{Chen Liang$^{1}$, Qian Wang$^{1}$, Andy Xu$^{1}$, Daniel Rakita$^{1}$
\thanks{$^{1}$All authors are in the Department of Computer Science, Yale University,
        New Haven, CT 06520, USA.
        {\tt\small daniel.rakita@yale.edu}}%
\thanks{This work was supported by Office of Naval Research award N00014-24-1-2124}
}

\begin{document}

\maketitle
\thispagestyle{empty}
\pagestyle{empty}


\begin{abstract}
The Rust programming language is an attractive choice for robotics and related fields, offering highly efficient and memory-safe code. However, a key limitation preventing its broader adoption in these domains is the lack of high-quality, well-supported Automatic Differentiation (AD)—a fundamental technique that enables convenient derivative computation by systematically accumulating data during function evaluation.  In this work, we introduce ad-trait, a new Rust-based AD library. Our implementation overloads Rust's standard floating-point type with a flexible trait that can efficiently accumulate necessary information for derivative computation. The library supports both forward-mode and reverse-mode automatic differentiation, making it the first operator-overloading AD implementation in Rust to offer both options. Additionally, ad-trait leverages Rust’s performance-oriented features, such as Single Instruction, Multiple Data acceleration in forward-mode AD, to enhance efficiency.  Through benchmarking experiments, we show that our library is among the fastest AD implementations across several programming languages for computing derivatives.  Moreover, it is already integrated into a Rust-based robotics library, where we showcase its ability to facilitate fast optimization procedures. We conclude with a discussion of the limitations and broader implications of our work.
\end{abstract}





\section{Introduction}
\label{sec:introduction}

The Rust programming language is becoming increasingly popular in research and industry.  In many ways, it is an attractive choice for robotics and related fields, offering highly efficient, memory-safe, thread-safe, and reliable code while also supporting deployment on the web and embedded devices.  However, a key limitation preventing its broader adoption in these domains is its lack of support for easily and efficiently computing derivatives of functions.

Derivatives are fundamental to numerical optimization, guiding iterative processes toward a function's local or global minimum. While derivatives can theoretically be computed analytically, often allowing for fast evaluation, explicitly deriving them for large or complex functions is typically impractical. In contrast, Automatic Differentiation (AD) provides a systematic approach to computing derivatives by transparently accumulating information during function evaluation, eliminating the need for manual differentiation.


In this paper, we present ad-trait, a new automatic differentiation library implemented in Rust. At its core is a flexible trait called \texttt{AD} which overloads standard floating point operations. Types implementing this trait behave like regular floating-point numbers while optionally tracking derivative information under the hood. A key innovation is our implementation of the \texttt{AD} trait for Rust's native \texttt{f64} and \texttt{f32} types, enabling code to switch between standard floating-point operations and derivative computation with minimal performance overhead. This design allows for nearly zero-cost abstraction when transitioning between regular and AD-enabled computations. 

The library offers both forward-mode and reverse-mode automatic differentiation, making it the first operator-overloading AD implementation in Rust to offer both.  Importantly, types that implement the \texttt{AD} trait must also implement the already existing Rust traits \texttt{RealField} and \texttt{ComplexField}, meaning these types can automatically integrate into popular Rust-based numerical computing and linear algebra libraries such as Nalgbra and ndarray.  Additionally, ad-trait leverages Rust’s performance-oriented features, such as Single Instruction, Multiple Data (SIMD) acceleration in forward-mode AD, to enhance efficiency.


Through extensive benchmarking experiments, we show that our library is among the fastest AD implementations across several programming languages for computing derivatives.  These experiments show that the forward or reverse-mode options in our library scale well with number of function outputs or inputs, respectively.  Moreover, our library is already integrated into a Rust-based robotics library called apollo-rust, where we showcase its ability to facilitate fast optimization procedures. 


We provide open-source code for ad-trait\footnote{\href{https://github.com/djrakita/ad_trait}{https://github.com/djrakita/ad\_trait}} and the apollo-rust\footnote{\href{https://github.com/Apollo-Lab-Yale/apollo-rust}{https://github.com/Apollo-Lab-Yale/apollo-rust}} library that incoporates this AD library.
\section{Background}
\label{sec:bacground}

\subsection{Automatic Differentiation Primer}

The mathematical object we are trying to compute is the derivative object of a computable and differentiable function $f: \mathbb{R}^n \rightarrow \mathbb{R}^m$ at a given input $\mathbf{x}_k \in \mathbb{R}^n$, denoted as $\frac{\partial f}{\partial \mathbf{x}}\big|_{\mathbf{x}_k}$.  This derivative will be an $m \times n$ matrix, i.e., $\frac{\partial f}{\partial \mathbf{x}}\big|_{\mathbf{x}_k} \in \mathbb{R}^{m \times n}$.  This matrix is commonly referred to as a \textit{Jacobian}, or specifically as a \textit{gradient} when $m = 1$ \cite{stewart2012calculus}.

A common strategy for computing derivatives involves introducing small perturbations in the input or output space surrounding the derivative and incrementally constructing the derivative matrix by analyzing the local behavior exhibited by the derivative in response to these perturbations \citep{griewank2008evaluating}.    

Specifically, perturbing the derivative in the input space looks like the following: $\frac{\partial f}{\partial \mathbf{x}}\big|_{\mathbf{x}_k} \ \Delta \mathbf{x} = \Delta \mathbf{f}$.  The $\Delta \mathbf{x} \in \mathbb{R}^n$ object here is commonly called a \textit{tangent}, and the resulting $\Delta \mathbf{f}$ is known as the \textit{Jacobian-vector product} (JVP) or \textit{directional derivative} \cite{baydin2018automatic, griewank2008evaluating}.  

Conversely, perturbing the derivative in the output space looks like the following: $\Delta \mathbf{f}^\top \ \frac{\partial f}{\partial \mathbf{x}}\big|_{\mathbf{x}_k} = \Delta \mathbf{x}^\top$.  The $\Delta \mathbf{f}^\top \in \mathbb{R}^{1 \times m}$ object here is called an \textit{adjoint}, and the result $\Delta \mathbf{x}^\top$ is known as the \textit{vector-Jacobian product} (VJP) \cite{baydin2018automatic, griewank2008evaluating}.

Note that the concepts of JVPs and VJPs now give a clear strategy for isolating the whole derivative matrix.  For instance, using ``one-hot'' vectors for tangents or adjoints, i.e., vectors where only the $i$-th element is $1$ and all others are $0$, can effectively capture the $i$-th column or $i$-th row of the derivative matrix, respectively \cite{griewank2008evaluating}.  Thus, the derivative matrix can be fully recovered using $n$ JVPs or $m$ VJPs.  

The strategy outlined above precisely defines Automatic Differentiation.  When tangent information is propagated alongside standard forward computations, the method is known as Forward-mode AD. Conversely, when adjoint information is propagated in reverse, from the function's output back to its input, the method is called Reverse-mode AD. Reverse-mode AD involves constructing a Computation Graph (or Wengert List \cite{wengert1964simple}) that logs primitive operations through the function during the forward pass and then performing a reverse pass over this graph to backpropagate adjoint information.

\subsection{Related Works}  
Numerous general Automatic Differentiation implementations have been developed, which can be broadly classified into three main categories: (1) operator overloading, (2) source code transformation, and (3) tensor-based. 

Operator overloading replaces standard arithmetic operations on floating-point values with specialized functions that track derivatives throughout computation. This approach is particularly effective in compiled languages, such as C++ and Julia, which support compile-time polymorphism or dynamic dispatch, enabling flexible overloading without runtime overhead. For example, AutoDiff \cite{autodiff} and Eigen's AutoDiffScalar \cite{eigenweb} in C++ leverage dual numbers for forward-mode AD, seamlessly integrating with standard numerical computations. Additionally, AutoDiff supports reverse-mode AD, allowing efficient derivative computation for functions with many inputs and a few outputs. Similarly, ForwardDiff.jl \cite{RevelsLubinPapamarkou2016} and ReverseDiff.jl \cite{reverseDiff} in Julia use operator overloading for forward-mode and reverse-mode AD, respectively, taking advantage of Julia’s Just-In-Time (JIT) compilation and multiple dispatch for high-performance AD. 

Our ad-trait library follows an operator overloading approach similar to AutoDiff \cite{autodiff} in C++, supporting both forward-mode and reverse-mode AD. Our goal is to bring this type of high-performance, flexible AD implementation to pure Rust while leveraging Rust's trait-based abstractions and seamless SIMD integration to further enhance efficiency and usability.

Source code transformation is an AD approach where the original program is statically analyzed and rewritten to generate derivative computations automatically \cite{margossian2019review}. This method modifies the Intermediate Representation (IR) or Abstract Syntax Tree (AST) to propagate derivative information efficiently.  It is commonly used in languages that support meta-programming or compiler-level transformations, such as Zygote.jl \cite{zygote2018} in Julia, Tapenade \cite{hascoet2013tapenade} for Fortran and C, and Enzyme \cite{juliaEnzyme} for LLVM-based languages. 


In many scenarios, source code transformation can be more efficient than operator overloading, as it avoids the runtime overhead of dynamically tracking computations. However, at the time of writing, there is no well-supported source code transformation AD implementation in Rust. While Enzyme can work in Rust since Rust compiles via LLVM, at the time of writing, the Rust wrapper for Enzyme is currently unsupported. 


Tensor-based AD operates on structured arrays rather than individual scalar operations. These libraries leverage highly optimized tensor computation back-ends to perform differentiation efficiently, making them well-suited for deep learning and large-scale scientific computing. Unlike operator overloading or source code transformation, tensor-based AD frameworks, such as PyTorch \cite{paszke2019pytorch}, JAX \cite{jax2018github}, and TensorFlow, differentiate entire tensor operations instead of performing element-wise computations. This enables them to leverage GPU acceleration, parallelism, and JIT compilation, leading to high-performance derivative computation. 


While tensor-based AD is often faster than source code transformation for large-scale linear algebra operations, it is less efficient or flexible for scalar-heavy functions. While scalars can technically be represented as $1\times1$ ``tensors'', the overhead of managing their dimensions and structure is significantly higher than using a simple scalar type. Rust features well-supported tensor-based AD libraries such as Burn\footnote{\href{https://github.com/tracel-ai/burn}{https://github.com/tracel-ai/burn}} and Candle\footnote{\href{https://github.com/huggingface/candle}{https://github.com/huggingface/candle}} . However, these libraries are designed specifically for deep learning, only support reverse-mode AD, and are not optimized for scalar-heavy computations.
\section{Implementation Overview}
\label{sec:implmentation_overivew}

In this section, we overview our AD library and highlight the key design choices that leverage Rust's capabilities.  

Our implementation heavily relies on a Rust feature called \textit{traits}.  Rust traits define shared behavior across multiple types by specifying a set of functions that those types must implement. They enable polymorphism and code reuse, allowing different types to be treated uniformly as long as they implement the same trait. Traits can also enforce dependencies by requiring that a type implements other traits. Furthermore, they serve as generic constraints, ensuring that functions or structs operate only on types that exhibit the required functions and behavior. 


\subsection{AD Trait}


In the ad-trait library, we introduce a new trait called \texttt{AD}. Types implementing the \texttt{AD} trait must support all standard floating-point operations, including addition, subtraction, multiplication, division, sine, cosine, etc., as well as comparison operators such as greater than, less than, and equality checks. Additionally, they must also implement Rust's existing \texttt{RealField} and \texttt{ComplexField} traits, allowing seamless integration with popular Rust-based numerical computing and linear algebra libraries such as Nalgebra\footnote{\href{https://nalgebra.org/}{https://nalgebra.org/}} and NDarray\footnote{\href{https://docs.rs/ndarray/latest/ndarray/}{https://docs.rs/ndarray/latest/ndarray/}}.  For a complete list of requirements defined by the \texttt{AD} trait, please refer to our source code.


In Rust, any numerical computing code can seamlessly integrate with the \texttt{AD} trait by replacing standard floating-point values in functions and structs with generic types that implement \texttt{AD}.  For instance, the code snippet below shows the simple conversion of a standard function to a function that utilizes the \texttt{AD} trait.

\begin{figure}[h!]
\vspace{-0.45cm}
\centering
\begin{minted}[frame=lines, fontsize=\scriptsize, linenos, bgcolor=bg, numbersep=1pt]{rust}
// non-AD example
fn foo(a: f64, b: f64) -> f64 { a + b } 
// AD example
fn foo_ad<A: AD>(a: A, b: A) -> A { a + b }
\end{minted}
\label{fig:preprocessor_snippet}
\vspace{-1.25cm}
\end{figure}

\subsection{Types that implement AD Trait}

The ad-trait library provides implementations of the \texttt{AD} trait for 18 Rust types, as seen in be seen in Table \ref{tab:types}.  The standard floating-point types, \texttt{f32} and \texttt{f64}, enable functions and structs designed for a generic \texttt{AD} type to seamlessly support raw floating-point evaluation with minimal overhead, effectively providing a near zero-cost abstraction.  

\begin{table}[t!]
    \centering
    \renewcommand{\arraystretch}{1.5}
    \begin{tabular}{|>{\centering\arraybackslash}m{0.25\linewidth}|>{\centering\arraybackslash}m{0.66\linewidth}|}
        \hline
           & \textit{Provided implementations of} \texttt{AD} \textit{trait} \\ \hline
        \textbf{Standard floating point types} & \texttt{f32}, \texttt{f64} \\ \hline
        \textbf{Forward AD types} &  \texttt{adf\_f32x1}, \texttt{adf\_f32x2}, \texttt{adf\_f32x4}, \texttt{adf\_f32x8}, \texttt{adf\_f32x16}, \texttt{adf\_f32x32}, \texttt{adf\_f32x64}, \texttt{adf\_f64x1}, \texttt{adf\_f64x2}, \texttt{adf\_f64x4}, \texttt{adf\_f64x8}, \texttt{adf\_f64x16}, \texttt{adf\_f64x32}, \texttt{adf\_f64x64}, \texttt{adfn<N>} \\ \hline
        \textbf{Reverse AD types} & \texttt{adr} \\ \hline
    \end{tabular}
    \caption{ Types that implement \texttt{AD} trait }
    \label{tab:types}
    \vspace{-5pt}
\end{table}  

The forward AD types are all structs that have two fields: \texttt{value} and \texttt{tangent}.  The \texttt{value} field is always a standard \texttt{f64} number, while the \texttt{tangent} field contains a direction on this variable that should be used to generate Jacobian-vector products (JVPs) through downstream computations.  


In standard forward-mode AD, the tangent on a certain variable is just a single number, meaning $n$ passes through the function will be needed to isolate all $n$ columns of the derivative matrix.  However, in ad-trait, the tangent field can hold \textit{multiple} values that can ideally update simultaneously utilizing Rust's convenient Single Instruction, Multiple Data (SIMD) features.  For instance, the \texttt{adf\_f32x16} forward AD struct looks like the following:

\begin{figure}[h!]
\vspace{-0.5cm}
\centering
\begin{minted}[frame=lines, fontsize=\scriptsize, bgcolor=bg, linenos, numbersep=1pt]{rust}
pub struct adf_f32x16 { value: f64, tangent: f32x16 }
\end{minted}
\label{fig:adf_type1}
\vspace{-0.9cm}
\end{figure}

\noindent where \texttt{f32x16} is a native type in the experimental, nightly version of the Rust standard library.\footnote{\href{https://doc.rust-lang.org/std/simd/index.html}{https://doc.rust-lang.org/std/simd/index.html}}  If the hardware supports 16-way 32-bit floating-point arithmetic, the \texttt{f32x16} type will update all channels simultaneously.  Thus, in theory, this type can enable forward AD that requires 16$\times$ fewer calls to a function since many tangent channels can be run in parallel.  This same concept applies for all similar types (\texttt{adf\_f32x2}, \texttt{adf\_f32x4}, etc.).

The \texttt{adfn} forward AD type differs in that the length of the tangent can be set manually at compile time:

\begin{figure}[h!]
\vspace{-0.5cm}
\centering
\begin{minted}[frame=lines, fontsize=\scriptsize, bgcolor=bg, linenos, numbersep=1pt]{rust}
pub struct adfn<const N: usize> 
    { value: f64, tangent: [f64; N] }
\end{minted}
\label{fig:adf_type2}
\vspace{-1cm}
\end{figure}

\noindent Here, instead of relying on explicit SIMD types, the Rust compiler can automatically optimize performance for any given \texttt{N} under the hood.

Lastly, \texttt{adr} is the sole reverse-mode AD type. In this approach, all overloaded functions in the \texttt{AD} trait log their operation type and parent operands into a computation graph. This graph is stored in heap memory and is pre-allocated whenever possible to optimize memory writes. 


\subsection{Differentiable Function and Derivative Method Traits}

The ad-trait library provides a trait called \texttt{DifferentiableFunctionTrait} that encapsulates a computable, differentiable function:

\begin{figure}[h!]
\vspace{-0.5cm}
\centering
\begin{minted}[frame=lines, fontsize=\scriptsize, bgcolor=bg, linenos, numbersep=1pt]{rust}
pub trait DifferentiableFunctionTrait<T: AD> {
    fn call(&self, inputs: &[T]) -> Vec<T>;
    fn num_inputs(&self) -> usize;
    fn num_outputs(&self) -> usize;
}
\end{minted}
\label{fig:differentiable_function}
\vspace{-1cm}
\end{figure}

At a high level, any type that implements this trait will be able to be differentiated through ad-trait. 
Additionally, the library includes the \texttt{DerivativeMethodTrait}, which defines various methods for computing derivatives:   

\begin{figure}[h!]
\vspace{-0.5cm}
\centering
\begin{minted}[frame=lines, fontsize=\scriptsize, bgcolor=bg, linenos, numbersep=1pt]{rust}
pub trait DerivativeMethodTrait: Clone {
   type T: AD;

   fn derivative<D: DifferentiableFunctionTrait<Self::T>>
    (&self, inputs: &[f64], function: &D) -> 
        (Vec<f64>, DMatrix<f64>);
}
\end{minted}
\label{fig:derivative_method}
\vspace{-1cm}
\end{figure}

This trait requires two components: a type that implements \texttt{AD} and a derivative function that returns the function output and derivative matrix for a function at given inputs.  

Our library currently provides four standard implementations of the \texttt{DerivativeMethodTrait}: (1) \texttt{FiniteDifferencing} (where \texttt{T} is \texttt{f64}); (2) \texttt{ReverseAD} (where \texttt{T} is \texttt{adr}); (3) \texttt{ForwardAD} (where \texttt{T} is $\texttt{adfn<1>}$); and (4) \texttt{ForwardADMulti} (where \texttt{T} can be any forward AD type from Table \ref{tab:types}).   

\subsection{Differentiable Block Struct}

The traits outlined above provide all the essential components for seamlessly computing function values and derivatives for any differentiable function in \texttt{ad-trait}: (1) An instantiation of a type that implements \texttt{DerivativeMethodTrait} to handle differentiation; (2) An instantiation of a type implementing \texttt{DifferentiableFunctionTrait} with an \texttt{AD} type compatible with the chosen derivative method; and (3) An instantiation of the same function using \texttt{f64} as the \texttt{AD} type for standard function evaluation.

To streamline this setup, our library offers a convenience struct called \texttt{DifferentiableBlock}, which bundles these three components. This struct provides user-friendly methods such as \texttt{call} and \texttt{derivative}, enabling straightforward function evaluation and differentiation.

\subsection{Integration with Robotics Library}

We have integrated \texttt{ad-trait} with our Rust-based robotics library, \texttt{apollo-rust}, making all its functionality fully compatible with \texttt{ad-trait}. This is achieved by replacing standard floating-point values with \texttt{AD} trait generic types throughout the library. As a result, this integration enables seamless differentiation of various essential subroutines, including spatial computations using Lie Group and Lie Algebra concepts, proximity functions via the Gilbert–Johnson–Keerthi (GJK) \cite{ericson2004} algorithm, spline-based interpolations such as Bézier and B-splines \cite{Shirley2009}, as well as forward and inverse kinematics functions, among others.  Additionally, apollo-rust supports its own optimization solvers including BFGS, L-BFGS \cite{nocedal1999numerical}, etc., meaning the \texttt{AD} trait overloaded versions of these solvers can support fully differentiable optimization.  
\section{Evaluation 1: Benchmark Function}
\label{sec:evaluation1}

In Evaluation 1, we compare ad-trait to several other Automatic Differentiation libraries on a benchmark function.

\subsection{Procedure}
\label{sec:procedure}

Evaluation 1 follows a three step procedure: (1) The benchmark function shown in Algorithm \ref{alg:benchmark} in initialized with given parameters $n$, $m$, and $o$.  This function will remain fixed and deterministic through the remaining steps; (2) An ordered set of $w$ random states of dimensionality $n$ are generated; and (3) For all conditions, derivatives of the benchmark function are computed in order on the $w$ inputs in the set from Step 2.  This ordered set of inputs is kept fixed for all conditions.  We record and report on the average runtime (in seconds) of derivative computation through the sequence of $w$ inputs.

The benchmark function used in this experiment is a randomly generated composition of sine and cosine functions. This function was designed to be highly parameterizable, allowing for any number of inputs ($n$), outputs ($m$), and operations per output ($o$). Sine and cosine were selected for their smooth derivatives and composability, given their infinite domain and bounded range.  Moreover, numerous subroutines in robotics and related fields involve many compositions of sine and cosine functions, making this function a reasonable analogue of these processes.

All experiments in Evaluation 1 were run on a Desktop computer with an Intel i7-14700 5.4GHz processor, 32 GB of RAM, and NVIDIA GeForce RTX 4080 GPU with CUDA enabled.  

\algBenchmark

\subsection{Conditions}
\label{sec:conditions}

Evaluation 1 compares fourteen conditions, six using forward mode AD and eight using reverse mode AD.  

The forward mode AD conditions are (1) JAX \cite{jax2018github} in Python, JIT-compiled on GPU; (2) ForwardDiff.jl \cite{RevelsLubinPapamarkou2016} in Julia; (3) Enzyme.jl \cite{juliaEnzyme} in Julia; (4) AutoDiff \cite{autodiff} in C++; (5) ad-trait (ours) in Rust; and (6) ad-trait (ours) in Rust that includes SIMD acceleration, using the \texttt{ForwardADMulti} derivative method with \texttt{adfn<N>} type where \texttt{N} is always set to be the number of function inputs.

The reverse mode AD conditions are (1) Pytorch in Python; (2) JAX \cite{jax2018github} in Python, JIT-compiled on GPU; (3) ReverseDiff.jl \cite{reverseDiff} in Julia; (4) Enzyme.jl \cite{juliaEnzyme} in Julia; (5) Zygote.jl \cite{zygote2018} in Julia; (6) AutoDiff \cite{autodiff} in C++; (7) Burn in Rust with ndarray backend; and (8) ad-trait (ours) in Rust.

All conditions were implemented in a best-effort fashion to achieve best possible performance. For instance, the JAX conditions used Just-In-Time (JIT) compilation onto a GPU, type hinting was used in Julia conditions, and heap allocations were avoided when possible in the C++ and Rust conditions. We provide a link to all implementations used in this experiment.\footnote{\href{https://github.com/chen-dylan-liang/AD-TRAIT}{https://github.com/chen-dylan-liang/AD-TRAIT}}

\subsection{Sub-experiment 1: Gradient Calculations}
\label{sec:subexp1}

In sub-experiment 1, we run the procedure outlined in \S\ref{sec:procedure} with parameters, $m=1$, $o=1000$, $w=100$, and $n=\{ 1, 50, 100, 150, ..., 1000 \}$.  Our goal in sub-experiment 1 is to observe how the different conditions scale as the number of function inputs $n$ grows while $m$ remains fixed at $1$.  In other words, the benchmark function here is a scalar function and its derivative is a $1 \times n$ row-vector gradient.

Results for Sub-experiment 1 are shown in Figure \ref{fig:se1} (top). As expected, reverse mode AD scales better than forward mode AD, and ad-trait ranks among the top performers in both categories. The default forward AD version of ad-trait closely matches AutoDiff in C++, but with SIMD and other compiler optimizations, it significantly outperforms AutoDiff and even surpasses Enzyme.jl for $700 \leq n \leq 1000$. In reverse mode AD, ad-trait again matches AutoDiff in C++, while Enzyme.jl remains the fastest.    

\subsection{Sub-experiment 2: Square Jacobian Calculations}
\label{sec:subexp2}

In sub-experiment 2, we run the procedure outlined in \S\ref{sec:procedure} with parameters, $n,m = \{ 1, 10, 20, 30, 40, 50 \}$, $o=1000$, and $w=100$.  Our goal in sub-experiment 2 is to observe how the different conditions scale as the number of function inputs $n$ and number of function outputs both grow.  Thus, the benchmark function here is a vector function with the same number of inputs and outputs, and its derivative is an $n \times n$ square Jacobian.

\begin{figure*}[t!]
\centering
\includegraphics[width=\textwidth]{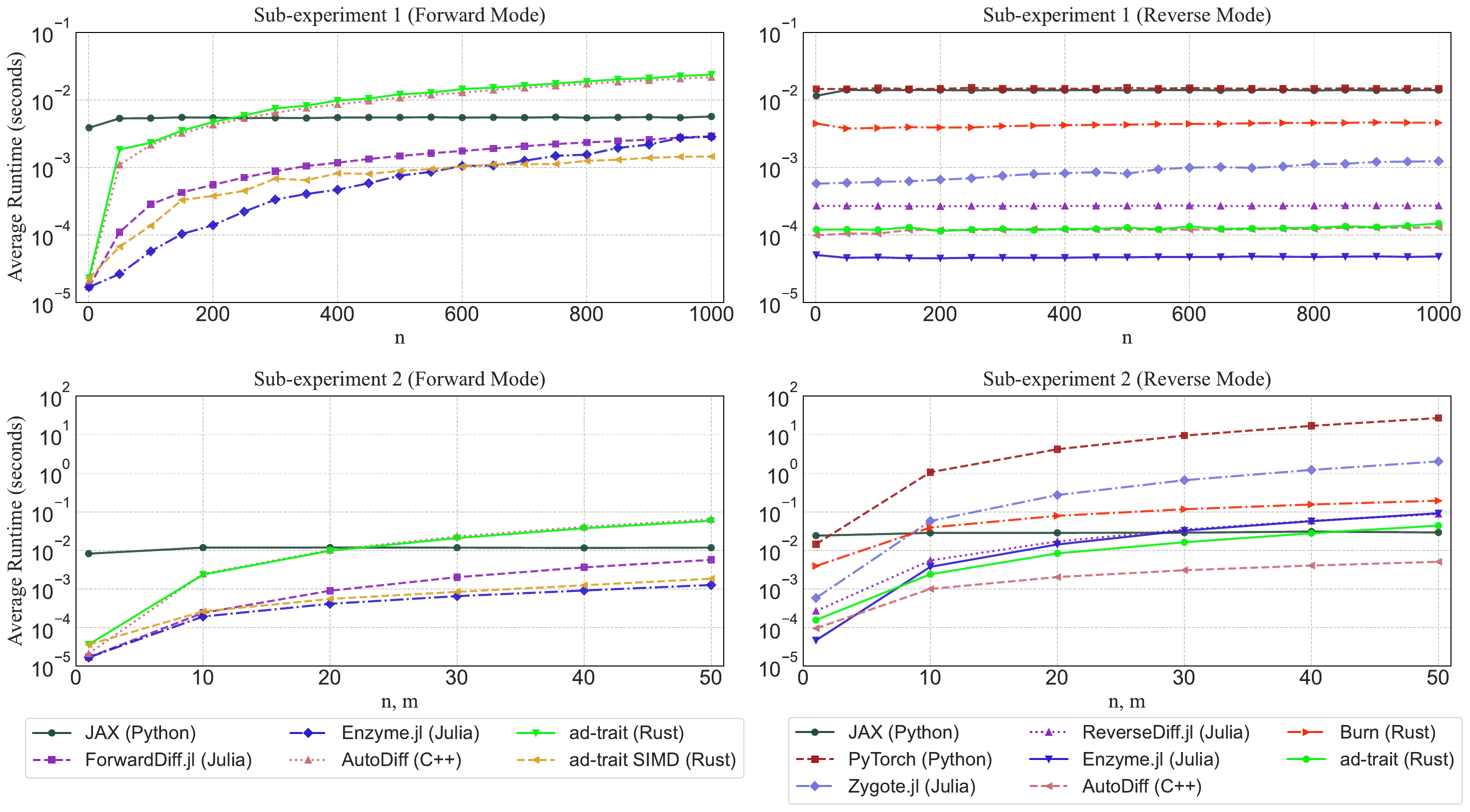}
\caption{ Results for Evaluation 1, Sub-experiment 1 (Top) and Sub-experiment 2 (Bottom).  Both Sub-experiments are split into forward mode AD conditions (left) and reverse mode AD conditions (right). }
\label{fig:se1}
\vspace{-0.2cm}
\end{figure*}

Results for Sub-experiment 2 are shown in Figure \ref{fig:se1} (bottom). While forward AD appears to outperform reverse AD, the trend is less distinct than in Sub-experiment 1, as both scale with the number of function inputs and outputs. In both cases, ad-trait remains among the top performers. Its default forward AD version closely matches AutoDiff in C++ but, with SIMD and other compiler optimizations, surpasses AutoDiff and matches Enzyme.jl. For reverse mode, AutoDiff in C++ is the fastest, with ad-trait generally ranking second, closely matching Enzyme.jl.  The ad-trait implementation was considerably faster on this benchmark than the tensor-based AD in Rust provided by Burn.

\begin{figure}[t!]
\centering
\includegraphics[width=\columnwidth]{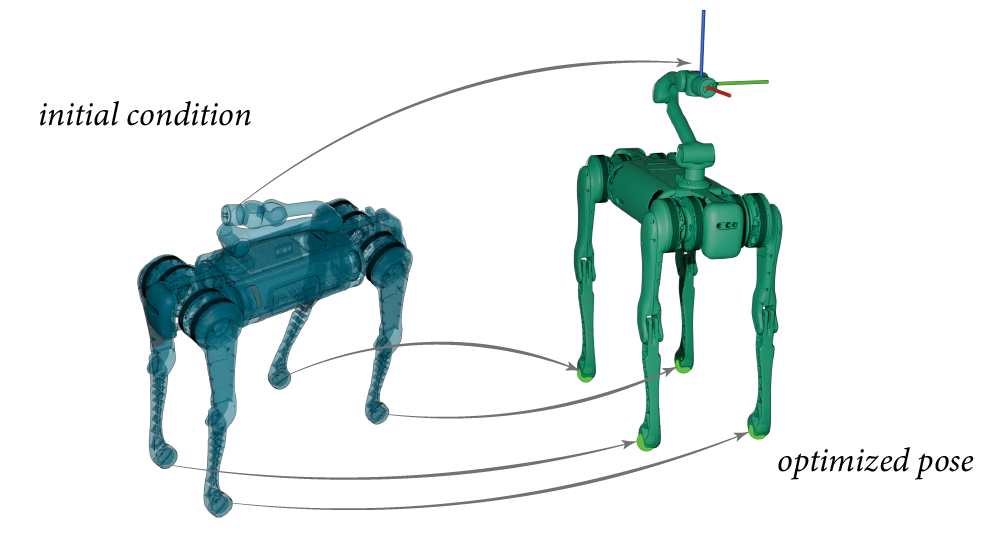}
\caption{ Evaluation 2 involves assessing performance in a robotic root-finding procedure, where the goal is to determine a robot pose that positions its feet and end-effector at predefined locations or orientations. }
\label{fig:robots}
\vspace{0.1cm}
\end{figure}

\section{Evaluation 2: Robotics Application}
\label{sec:evaluation2}

In Evaluation 2, we demonstrate that the derivative computation in ad-trait can work well in a robotics-based root-finding procedure.

\subsection{Procedure}

Evaluation 2 follows a three step procedure: (1) A robot state is sampled for a simulated Unitree B1 quadruped robot\footnote{\href{https://www.unitree.com/b1}{https://www.unitree.com/b1}} with a Z1 manipulator\footnote{\href{https://www.unitree.com/z1}{https://www.unitree.com/z1}} mounted on its back.  This robot has 24 degrees of freedom (including a floating base to account for mobility).  This sampled state, $\mathbf{x}_0 \in \mathbb{R}^{24}$, will be an initial condition for an optimization process; (2) A Jacobian pseudo-inverse method (Algorithm \ref{alg:root_finding}) is used to find a root for a constraint function (Algorithm \ref{alg:constraint_function}).  The constraint function has five outputs: four for specifying foot placements and one for specifying the end-effector pose for the manipulator mounted on the back.  Thus, the Jacobian of this constraint function is a $5 \times 24$ matrix; (3) Step 1--2 are run 500 times per condition.  We record and report on one metrics in Evaluation 2: the average runtime (in seconds) to converge on a robot configuration sufficiently close to the constraint surface.

All experiments in Evaluation 2 were run on a Desktop computer with an Intel i7-14700 5.4GHz processor with 32 GB of RAM.  


\algJacobianPseudoinverse
\algConstraintFunction

\subsection{Conditions}
The conditions in Evaluation 2 are all ad-trait derivative method options as these are integrated with the Rust-based robotics library apollo-rust.  Specifically, these are conditions 10--12 listed in \S\ref{sec:conditions}, as well as finite differencing implemented in ad-trait. 


\subsection{Results}

Results for Evaluation 2 are shown in Table \ref{tab:evaluation2} (with range values representing standard deviation). All ad-trait derivative methods successfully completed the root-finding procedure, which converged in approximately 300–500 iterations, depending on the initial state. Forward AD with SIMD optimization was the fastest. These results highlight ad-trait’s effectiveness in robotics applications, allowing seamless switching between derivative backends as single-parameter drop-in replacements.  

\begin{table}[t]
    \vspace{2pt}
    \centering
    \begin{tabular}{|>{\centering\arraybackslash}p{0.55\linewidth}|>{\centering\arraybackslash}p{0.35\linewidth}|}
        \hline
        \textit{Approach in ad-trait} & \textit{Average runtime (seconds $\pm$ standard deviation)} \\ \hline
        \textbf{Forward AD} & 0.035 $\pm$ 0.0021 \\ \hline
        \textbf{Forward AD (SIMD with \texttt{adfn<24>}) } & 0.017 $\pm$ 0.00094 \\ \hline
        \textbf{Reverse AD} & 0.098 $\pm$ 0.0054 \\ \hline
        \textbf{Finite Differencing}  & 0.022 $\pm$ 0.0012 \\ \hline
    \end{tabular}
    \caption{ Results for Sub-experiment 2 }
    \label{tab:evaluation2}
    \vspace{-1pt}
\end{table}
\section{Discussion}
\label{sec:discussion}

In this paper, we introduced ad-trait, a new automatic differentiation (AD) library implemented in Rust. Leveraging Rust’s trait system, we designed a flexible \texttt{AD} trait that enables operator overloading for standard floating-point operations. The ad-trait library supports both forward-mode and reverse-mode automatic differentiation, making it the first operator-overloading AD implementation in Rust to offer both. Additionally, it takes advantage of Rust’s performance-oriented features, such as Single Instruction, Multiple Data (SIMD) acceleration in forward-mode AD, to enhance computational efficiency.

Through our evaluations, we demonstrated that ad-trait ranks among the fastest AD implementations across multiple programming languages for computing derivatives. We further showcased its practical utility through a robotics-based optimization application. Through the rest of this section, we discuss the limitations and implications of our library.

\subsection{Limitations}

We identify several limitations in our implementation that open avenues for future research and extensions.  First, while developing new code with generic \texttt{AD} types is straightforward since they behave just like standard floating-point numbers, integrating \texttt{AD} types into existing code can be tedious. This process requires systematically replacing every floating-point value across functions and structs with a generic \texttt{AD} type, which can be cumbersome.  To address this, we aim to explore ways to automate the conversion process. Potential approaches include leveraging Rust macros for meta-programming or employing learning-based techniques, such as large language models (LLMs), to assist in refactoring code more efficiently.

Additionally, while operator overloading makes AD convenient to use, since overloaded types behave like standard floating-point values, it can be inefficient for certain reverse-mode AD computations. Specifically, every primitive operation in reverse-mode AD requires a heap memory write, which remains significantly slower than performing the raw computation, even with pre-allocation.  For example, in matrix-matrix multiplication, which involves a substantial number of multiplications and additions between individual elements, the overhead from repeated memory writes can severely degrade performance. In domains like machine learning, where vector, matrix, and tensor operations are prevalent, derivative computations are typically handled at the vector, matrix, or tensor level rather than at the individual variable level to mitigate these inefficiencies.

To address this limitation, we aim to extend ad-trait to support vector, matrix, and tensor-level derivatives, ensuring better performance in applications that rely heavily on structured mathematical operations. 

\subsection{Implications}

Through this work, we have demonstrated that ad-trait is a high-quality automatic differentiation library, implemented entirely in Rust. We hope this library will inspire further development in Rust for robotics, optimization, and related fields, fostering a new wave of implementations that prioritize memory safety, thread safety, portability, and reliability.

Moreover, the trait-based design of ad-trait offers ample opportunities for ongoing expansion. New implementations of the \texttt{AD} and \texttt{DerivativeMethodTrait} traits can be contributed by both us and the broader community, enabling continuous growth and innovation. We look forward to supporting and advancing this development going forward.    





\bibliographystyle{plainnat}

\bibliography{refs}


\end{document}